\documentclass{article}

\usepackage[final,nonatbib]{neurips_2024}
\makeatletter
\renewcommand{\@noticestring}{}  
\makeatother

\usepackage[utf8]{inputenc}
\usepackage[T1]{fontenc}
\usepackage{hyperref}
\usepackage{url}
\usepackage{booktabs}
\usepackage{amsfonts}
\usepackage{amsmath}
\usepackage{amssymb}
\usepackage{nicefrac}
\usepackage{microtype}
\usepackage{graphicx}
\usepackage{xcolor}
\usepackage{subcaption}
\usepackage{multirow}
\usepackage{natbib}

\newcommand{\nats}{\,\text{nats}}

\title{Marginals Before Conditionals:\\Staged Disambiguation in Gradient-Trained Transformers}

\author{
  Mihir Sahasrabudhe \\
  School of Information Sciences \\
  University of Illinois Urbana-Champaign \\
  \texttt{mihirrs2@illinois.edu}
}

\begin{document}

\maketitle

\begin{abstract}
We construct a minimal task that isolates conditional learning in neural networks: a surjective map with $K$-fold ambiguity, resolved by a selector token $z$, so $H(A \mid B) = \log K$ while $H(A \mid B, z) = 0$.
The model learns the marginal $P(A \mid B)$ first, producing a plateau at exactly $\log K$, before acquiring the full conditional in a sharp, collective transition.
The plateau has a clean decomposition: height $= \log K$ (set by ambiguity), duration $= f(D)$ (set by dataset size $D$, \emph{not} $K$).
Gradient noise stabilizes the marginal solution: higher learning rates monotonically slow the transition ($3.6\times$ across a $7\times$ $\eta$ range at fixed throughput), and batch-size reduction delays escape, consistent with an entropic force opposing departure from the low-gradient marginal.
Internally, a selector-routing head assembles during the plateau, leading the loss transition by ${\sim}50\%$ of the waiting time.
This is the Type~2 directional asymmetry of~\citet{papadopoulos2024arrows}, measured dynamically: we track the excess risk from $\log K$ to zero and characterize what stabilizes it, what triggers its collapse, and how long it takes.
\end{abstract}

\noindent\textbf{Code, data-generation scripts, and trained checkpoints:} \url{https://github.com/mihirrs/synass-lens}

\section{Introduction}
\label{sec:intro}

Neural networks learn structured tasks in stages.
Delayed generalization can follow memorization by orders of magnitude in training steps~\citep{power2022grokking}; hidden progress accumulates long before external metrics move~\citep{barak2022hidden}; and models routinely fail to exploit information present in their inputs~\citep{berglund2024reversal, kitouni2024factorization}.
Mechanistic studies of grokking have identified specific circuit-formation events underlying delayed generalization~\citep{nanda2023progress}, including Fourier-basis representations for modular arithmetic.
We study a related but distinct transition: from marginal to conditional prediction, rather than from memorization to generalization.

This connects to two lines of work on directional asymmetry.
\citet{papadopoulos2024arrows} showed that forward-trained language models outperform backward-trained ones (an ``arrow of time'') and identified a Type~2 asymmetry where the model \emph{can} represent the inverse but struggles to learn it.
The reversal curse~\citep{berglund2024reversal} showed that models trained on ``$A$ is $B$'' cannot infer ``$B$ is $A$.''
Both can be viewed through the lens of \emph{excess risk} in the inverse direction:
$\Delta_{X \to Y}(t) = \mathbb{E}[-\log q_\theta(Y \mid X)] - H(Y \mid X)$.
The arrows-of-time result shows these excess risks differ in aggregate; the reversal curse shows $\Delta_{B \to A}$ stays large for specific relations.
We measure $\Delta(t)$ as a function of training time and characterize its dynamics (plateau, snap, scaling, stabilization) in a controlled setting where every variable is isolated.

We construct a task where $\Delta(t) = \mathcal{L}(t)$ starts at exactly $\log K$ and drops to zero when the model learns to condition on a selector $z$.
Across hundreds of runs, the plateau duration depends on dataset size $D$, not disambiguation complexity $K$.
The marginal solution is stabilized by gradient noise (an entropic force in the sense of~\citet{ziyin2024sgd}), and the transition is collective: zero of 200 base groups are solved at $\tau/2$, then all snap simultaneously.

\paragraph{Contributions.}
(i)~A controlled task (a ``wind tunnel'' for studying conditional learning) with exact information-theoretic benchmarks (\S\ref{sec:task}).
(ii)~Plateau duration scales with dataset size $D$, not ambiguity $K$ (\S\ref{sec:duration}).
(iii)~Evidence for collective transition, entropic stabilization ($3.6\times$ LR effect, $1.8\times$ batch-size residual in tokens processed), and internal circuit formation (\S\ref{sec:transition}, \S\ref{sec:mechanism}).
(iv)~A directional asymmetry connecting to the reversal curse: unambiguous $A \to B$ is $1.7$--$4.4\times$ slower than structured $(B, z) \to A$ (\S\ref{sec:asymmetry}).
(v)~Seven falsified candidate mechanisms constraining the space of viable explanations (Appendix~\ref{app:falsifications}).

\section{Task and Apparatus}
\label{sec:task}

\paragraph{Task.}
A surjective map with $n_b$ base strings $B$ (6 characters from 36 alphanumeric symbols) and constant fiber size $K$: each $B$ maps to $K$ distinct targets $A$ (4 characters).
A selector $z$ (2 characters) indexes into the fiber, making $(B, z) \mapsto A$ one-to-one.
The model receives $[\texttt{BOS},\, B,\, \texttt{SEP},\, z,\, \texttt{SEP}]$ and predicts $A$ autoregressively.
Total dataset: $D = n_b \times K$ unique examples.
The task has exact entropy decomposition: $H(A \mid B) = \log K$, $H(A \mid B, z) = 0$.
A model ignoring $z$ achieves loss $\log K$; a model using $z$ achieves loss $0$.

\paragraph{Model.}
4-layer Transformer ($d{=}128$, 4 heads, $d_\text{mlp}{=}512$; ${\sim}600$K params), AdamW~\citep{loshchilov2019adamw}, batch size 128, $\eta = 10^{-3}$, cosine warmup 500 steps, via TransformerLens~\citep{nanda2022transformerlens}.

\paragraph{Diagnostics.}
The $z$-shuffle gap $\Delta_z = \mathcal{L}_{z\text{-shuffle}} - \mathcal{L}_\text{clean}$ detects when $z$ affects the output ($\Delta_z = 0$ during plateau; $\Delta_z \gg 0$ after).
It is computed at each evaluation step by running inference twice on the same batch: once with original $z$ tokens, once with $z$ tokens randomly permuted within the batch (preserving $z$ marginals but breaking the $(B, z) \to A$ mapping).
$\Delta_z$ is the difference in per-example cross-entropy loss, averaged over the batch.
Onset is defined as the first step where $\Delta_z > 0.1\nats$ for 3 consecutive evaluations.
The waiting time $\tau$ is when loss first drops below $50\%$ of $\log K$; this threshold is robust (exponent varies $< 0.1$ across thresholds $0.3$--$0.7$; Appendix~\ref{app:threshold}).

\paragraph{Seeds and replication.}
Unless otherwise noted, results are from single-seed runs (seed~42).
Where multi-seed data is available (phase-boundary experiments, 3 seeds per condition; Appendix~\ref{app:multiseed}), we report means and standard deviations.
Individual $\tau$ measurements have substantial seed variance (CV $30$--$70\%$ in phase-boundary runs), which is typical of stochastic processes near a bifurcation: the time to align with a shallow escape direction is inherently noisy.
Single-seed sweeps derive statistical power from consistency across many conditions (10 $D$~values, five batch sizes, four learning rates) rather than per-condition replication.

\section{The Phase Transition}
\label{sec:transition}

\subsection{The marginal plateau}

\begin{figure}[t]
  \centering
  \includegraphics[width=0.78\textwidth]{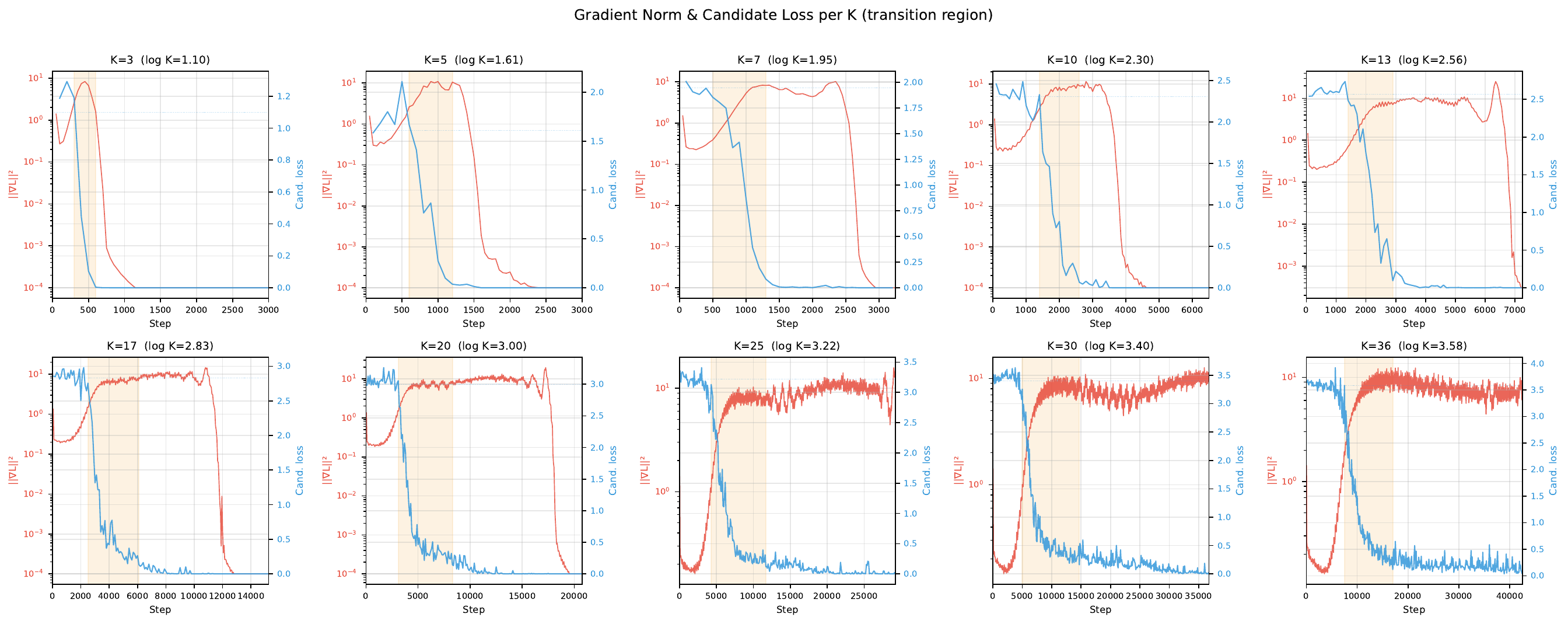}
  \caption{\textbf{Staged disambiguation.} Loss curves for $K \in \{5, 10, 20, 36\}$ at $n_b{=}1{,}000$. Dashed lines mark $\log K$. The model converges to the marginal solution within a few hundred steps, then plateaus until a sharp transition to near-zero loss.}
  \label{fig:main}
\end{figure}

Every run with $K > 1$ exhibits two regimes (Figure~\ref{fig:main}).
Loss drops to $\approx \log K\nats$ within a few hundred steps ($\Delta_z = 0$: the model ignores $z$), then plateaus for thousands of steps before a sharp transition to near-zero loss.
The plateau height tracks $\log K$ with ratio $1.01 \pm 0.02$ across $K \in \{3, \ldots, 36\}$: the model converges to the exact uniform-over-candidates solution, then stalls at the information-theoretic limit of that partial solution.

\subsection{Duration depends on dataset size, not ambiguity}
\label{sec:duration}

Since $D = n_b \times K$, varying $K$ at fixed $n_b$ changes both ambiguity and dataset size simultaneously.
To isolate the two factors, we hold $D$ constant by adjusting $n_b = D/K$.
At $D = 10{,}000$: $\tau \approx 1{,}600$ for all $K \in \{5, 10, 20, 36\}$ ($\delta = -0.004$, CI $[-0.04, 0.04]$).
At $D = 20{,}000$: $\tau \approx 3{,}950$ with $\delta = -0.04$ (CI $[-0.16, 0.04]$; Figure~\ref{fig:confound} in Appendix~\ref{app:confound}).
$K$ has no effect on waiting time once $D$ is controlled.
Note that holding $D$ constant while varying $K$ necessarily varies $n_b = D/K$ (e.g., at $D{=}10{,}000$: $K{=}5$ gives $n_b{=}2{,}000$ while $K{=}36$ gives $n_b{\approx}278$).
The flat $\tau$ across these conditions means neither $K$ nor $n_b$ matters independently---only their product $D$.

Since $K$ contributes nothing to $\tau$ beyond its role in determining $D$, the 10-point K-sweep at $n_b = 1{,}000$ (Appendix~\ref{app:ksweep}) is equivalently a 10-point D-sweep over $D \in \{3{,}000, \ldots, 36{,}000\}$, spanning a full decade.
The argument is two steps: (1)~the fixed-$D$ control establishes that $K$ has no independent effect on $\tau$; (2)~therefore every $\tau$ value in the K-sweep is attributable to $D = 1{,}000K$, not to $K$ per se.
Fitting $\tau$ versus $D$ yields $\tau \propto D^{1.19}$ ($R^2 = 0.98$, 95\% CI $[1.14, 1.31]$; Figure~\ref{fig:tau-vs-D}, Table~\ref{tab:ksweep}).
The fixed-$D$ control establishes $K$-independence; the K-sweep, reinterpreted as a D-sweep, provides the scaling law on 10 data points spanning a full decade.

\textbf{Central finding: plateau height depends on $K$ (how many candidates per group), but plateau duration depends on $D$ (how many examples the optimizer must process).}
The model must learn to route $z$ across all $D$ pairs; more pairs means more optimization work, regardless of how ambiguity is structured.
Multi-seed experiments (3 seeds per condition) confirm that $\tau$ increases with $D$ despite substantial seed-to-seed variability (CV $22$--$71\%$; Appendix~\ref{app:multiseed}).
The super-linear exponent remains unexplained.

\subsection{The collective snap}

Per-group evaluation at $K \in \{10, 20\}$ shows that the transition is collective, not incremental.
At $\tau/2$: zero of 200 sampled groups exceed $80\%$ accuracy (mean accuracy $15.4\%$ at $K{=}10$, $8.3\%$ at $K{=}20$).
At $\tau$, all groups snap together within a narrow window ($99\%$ of groups $\geq 80\%$ at $1.5\tau$; full data in Table~\ref{tab:collective}).
A shared circuit becomes operational for all groups simultaneously.

The transition width, defined as the step interval over which the fraction of groups exceeding $80\%$ accuracy rises from ${<}5\%$ to ${>}80\%$, spans approximately $0.5\tau$: at $K{=}10$, from step ${\sim}1{,}400$ to ${\sim}2{,}400$; at $K{=}20$, from ${\sim}3{,}500$ to ${\sim}5{,}200$ (Table~\ref{tab:collective}).
No groups solve before $0.5\tau$; the earliest solvers appear near $0.9\tau$, with the bulk following within $0.5\tau$ thereafter.
The transition is sharp relative to the plateau but occupies a measurable window, not a single step.
Under independent per-group learning with mean transition time $\tau$ and typical seed variance (CV ${\sim}30\%$), the expected transition window for $n_b{=}1{,}000$ groups would be negligible relative to $\tau$.
The observed $0.5\tau$ window suggests the transition propagates through the network over a measurable timescale, but is far narrower than what incremental group-by-group coverage would predict.

\subsection{Entropic stabilization}
\label{sec:entropic}

If the plateau were a local minimum, noise should facilitate escape.
The opposite holds, consistent with an \emph{entropic force}~\citep{ziyin2024sgd} stabilizing the marginal solution.
We distinguish manipulations that alter gradient noise \emph{without} changing the task from those that also degrade the conditional signal.

\paragraph{Task-preserving noise manipulations.}
Two manipulations change gradient noise while leaving the $(B, z) \to A$ mapping intact.

\begin{figure}[t]
  \centering
  \begin{subfigure}[t]{0.48\textwidth}
    \centering
    \includegraphics[width=\textwidth]{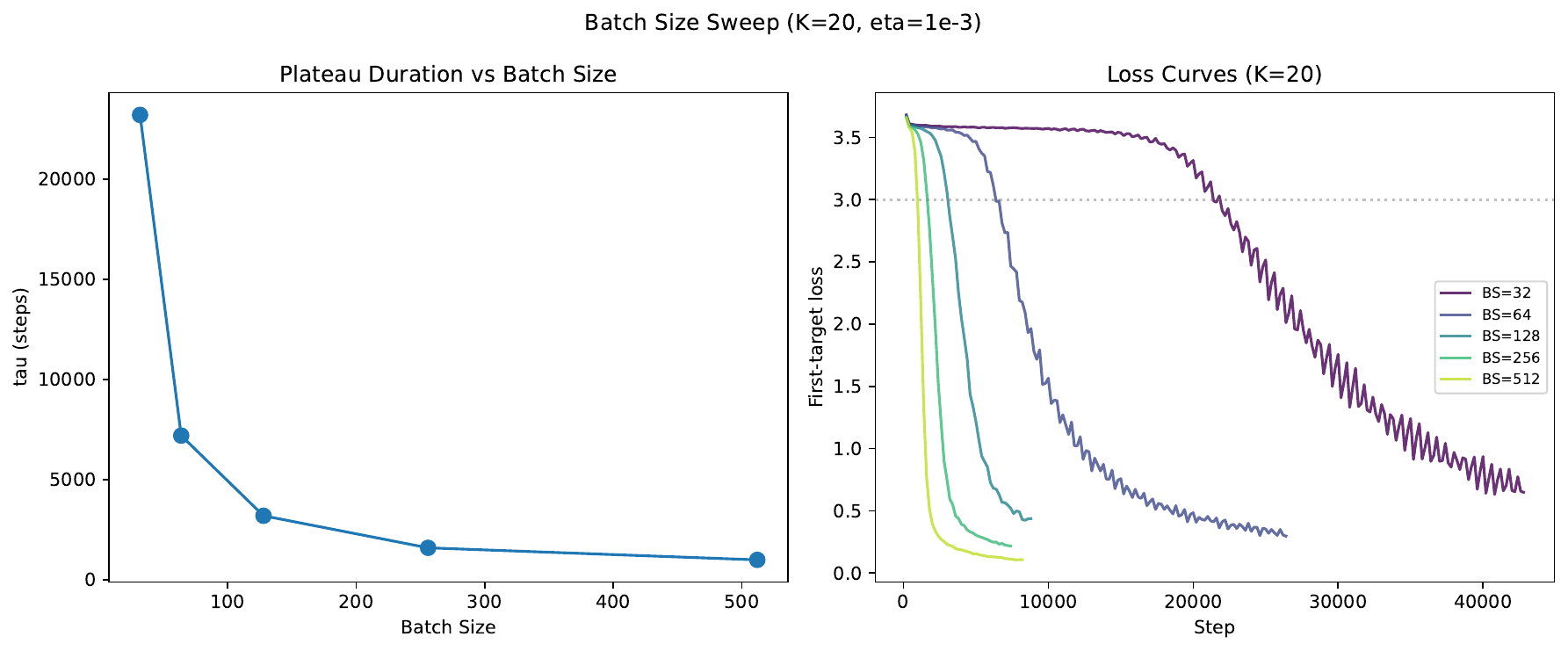}
    \caption{Batch-size sweep ($1.8\times$ in tokens; $23\times$ in steps is mostly throughput).}
    \label{fig:batch-sweep}
  \end{subfigure}
  \hfill
  \begin{subfigure}[t]{0.48\textwidth}
    \centering
    \includegraphics[width=\textwidth]{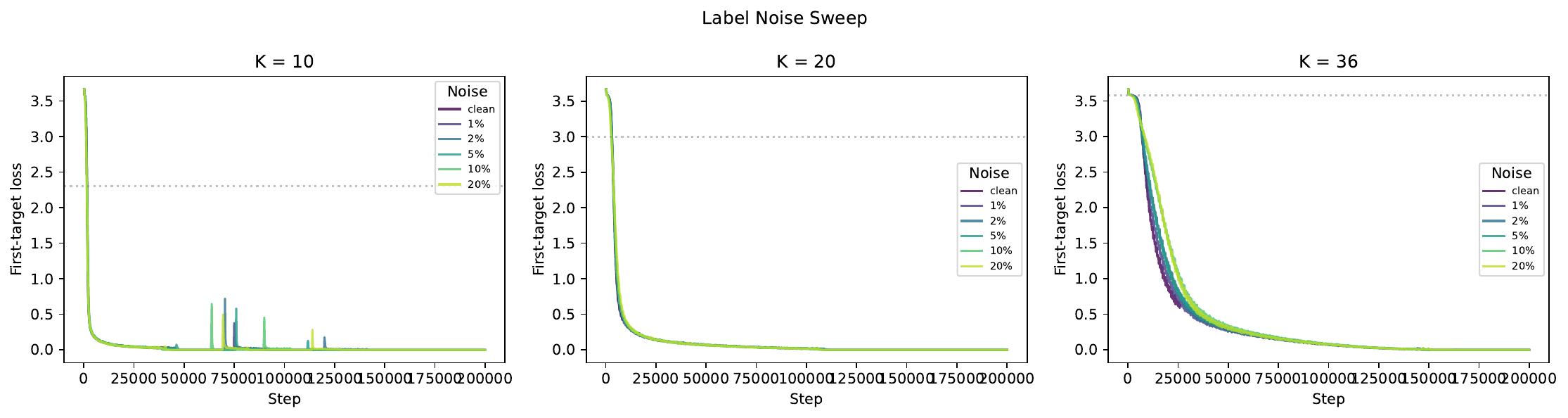}
    \caption{Label noise ($16\times$ in tokens at fixed $B{=}128$).}
    \label{fig:noise-sweep}
  \end{subfigure}
  \caption{\textbf{Noise delays escape.} (a)~Batch-size effect: a modest $1.8\times$ delay in tokens processed (the $23\times$ step-count ratio is predominantly throughput), consistent with entropic stabilization~\citep{ziyin2024sgd}. (b)~Label noise shows a $16\times$ delay in tokens, but also degrades the conditional signal (\S\ref{sec:entropic}); the LR sweep ($3.6\times$ at constant throughput) provides the cleanest evidence.}
  \label{fig:stabilization}
\end{figure}

\textbf{Batch size} (fixed $\eta{=}10^{-3}$).
Smaller batches increase per-step gradient noise without altering the task or learning rate~\citep{smith2018dont}.
At $K{=}20$, sweeping $B \in \{32, 64, 128, 256, 512\}$: the step-count ratio spans $23\times$, but smaller batches also reduce per-step throughput.
In the appropriate currency---tokens processed ($\tau_\text{tok} = \tau_\text{steps} \times B$)---the effect is a $1.8\times$ delay ($409$K--$742$K tokens; Table~\ref{tab:token-norm}, Figure~\ref{fig:stabilization}a).
This $1.8\times$ residual is modest but real and in the predicted direction: increased per-step noise delays escape even after controlling for data throughput.

\textbf{Learning rate} (fixed $B{=}128$).
Higher $\eta$ increases both step size and effective gradient noise.
$\tau$ increases monotonically with $\eta$: from $2{,}100$ steps ($269$K tokens) at $\eta = 3 \times 10^{-4}$ to $7{,}550$ steps ($966$K tokens) at $\eta = 2 \times 10^{-3}$; fails to converge at $\eta = 5 \times 10^{-3}$.
Because batch size is constant, the $\mathbf{3.6\times}$ token-normalized ratio is entirely attributable to the change in $\eta$, not throughput.
The monotonic direction (more noise $\to$ longer plateau, no U-shape) is inconsistent with barrier crossing regardless of the step-size confound.

\begin{table}[t]
\centering
\caption{\textbf{Batch-size sweep at $K{=}20$, $\eta{=}10^{-3}$: steps vs.\ tokens.} The $23\times$ step ratio is predominantly a throughput effect; the token-normalized residual is $1.8\times$.}
\label{tab:token-norm}
\vspace{4pt}
\begin{tabular}{@{}cccccc@{}}
\toprule
$B$ & $\tau_\text{steps}$ & $\tau_\text{tok}$ (K) & Step ratio & Token ratio \\
\midrule
32  & 23{,}200 & 742  & $23.2\times$ & $1.81\times$ \\
64  & 7{,}200  & 461  & $7.2\times$  & $1.12\times$ \\
128 & 3{,}200  & 410  & $3.2\times$  & $1.00\times$ \\
256 & 1{,}600  & 410  & $1.6\times$  & $1.00\times$ \\
512 & 1{,}000  & 512  & $1.0\times$  & $1.25\times$ \\
\bottomrule
\end{tabular}
\end{table}

\paragraph{Task-degrading manipulation: label noise.}
Within-fiber label noise at rate $p$ replaces the correct target with a random candidate from the same $B$-group.
This both increases gradient stochasticity \emph{and} reduces $I(A; z \mid B)$: at rate $p$, a fraction $p$ of training examples provide incorrect $z \to A$ mappings, making the conditional task itself harder.
$\tau_\text{tok}$ grows from $410$K ($p{=}0$) to $640$K ($p{=}0.1$) to $6{,}550$K ($p{=}0.2$), a $16\times$ ratio at fixed $B{=}128$ (Figure~\ref{fig:stabilization}b).
The sharp elbow between $p{=}0.1$ and $p{=}0.2$ ($10\times$ jump) likely reflects both noise-induced stabilization and task degradation; we cannot separate the two.
We include label noise for completeness but note that it is not a pure noise manipulation; the batch-size and learning-rate sweeps, which preserve the task structure, provide cleaner evidence for noise-induced stabilization.

\paragraph{Interpretation.}
The marginal solution has low gradient norm because $K$ competing directions cancel within each group; noise penalizes departure from this low-gradient state, acting as a restoring force.
The batch-size sweep shows the direction is correct (noise $\to$ slower escape) with a $1.8\times$ residual after throughput normalization.
The learning-rate sweep---the cleaner test, since batch size and thus per-step throughput are held constant---confirms the direction with a $3.6\times$ effect.
Together, both task-preserving manipulations are inconsistent with barrier crossing and consistent with entropic stabilization~\citep{ziyin2024sgd}.

\paragraph{Connection to SGD theory.}
\citet{wang2023noise} show that minibatch SGD noise preferentially projects onto high-curvature directions, causing escape along \emph{flat} rather than sharp directions.
In our setting, the marginal solution sits at a saddle with extreme anisotropy ($\lambda_\text{max} \approx 2.8$, $\lambda_\text{min} \approx -0.005$ at $K{=}20$; \S\ref{sec:mechanism}): the dominant curvature does not lead to escape, while the shallow $\lambda_\text{min}$ direction does.
Noise-curvature alignment implies that stochastic perturbations predominantly excite the non-escape direction, and increasing noise amplitude strengthens this misalignment.

First-exit-time analyses~\citep{nguyen2019exit} predict that exit times from metastable states depend on both local geometry and the noise structure.
In our anisotropic saddle, the relevant escape is along a direction ${\sim}500\times$ shallower than the dominant curvature, making exit times sensitive to the noise distribution.
Adaptive noise injection~\citep{gong2025adaptive} shows that tuning noise tails to the landscape accelerates escape from sharp regions; the stabilization we observe is plausibly the mirror image, where untuned stochastic noise stabilizes flat regions by failing to project onto the escape direction.

Stronger causal tests of this mechanism would hold the task fixed while manipulating noise geometry directly: adding isotropic Gaussian noise to the gradient (isolating stochasticity from task structure), varying dropout rate (changing effective noise without altering inputs), or using gradient accumulation to decouple batch noise from per-step throughput.
Measuring the empirical gradient covariance and its alignment with Hessian eigenvectors during the plateau would provide a definitive test of the noise-curvature account.
We leave these to future work.

\section{Mechanism}
\label{sec:mechanism}

\begin{figure}[t]
  \centering
  \begin{subfigure}[t]{0.48\textwidth}
    \centering
    \includegraphics[width=\textwidth]{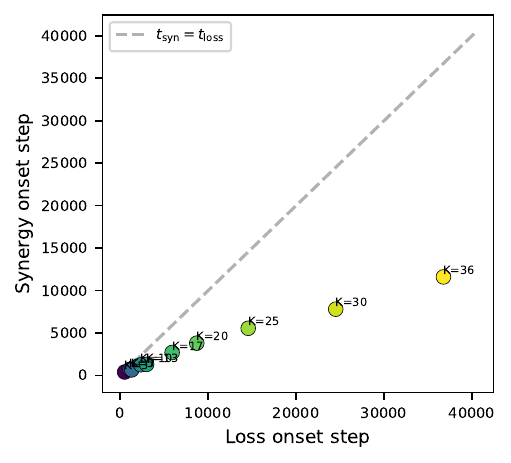}
    \caption{$\Delta_z$ onset leads loss onset across all $K$.}
    \label{fig:synergy-leads}
  \end{subfigure}
  \hfill
  \begin{subfigure}[t]{0.48\textwidth}
    \centering
    \includegraphics[width=\textwidth]{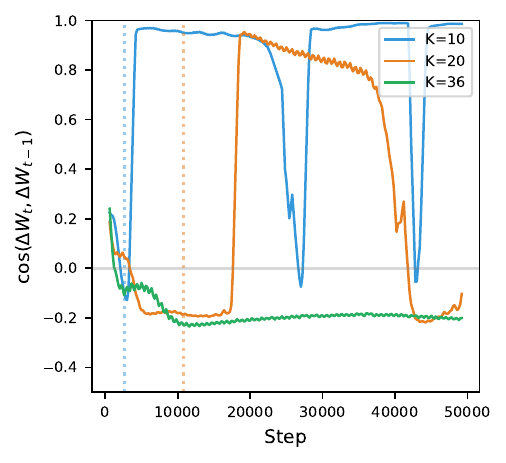}
    \caption{Direction consistency: ${\approx}0$ on plateau, ${\sim}0.8$ at transition.}
    \label{fig:direction-consistency}
  \end{subfigure}
  \caption{\textbf{Internal precursors.} (a)~$z$-dependence leads loss across all $K$. (b)~Transition is onset of \emph{coherent} descent.}
  \label{fig:cascade}
\end{figure}

\paragraph{Internal cascade.}
Across all $K$ values, $\Delta_z$ onset consistently precedes loss onset, with mean lead fraction $0.52 \pm 0.14$ of $\tau$ (Figure~\ref{fig:cascade}a).
At $K{=}20$, the aggregate $\Delta_z$ exceeds detection at step ${\sim}2{,}750$, well before loss drops at step ${\sim}8{,}700$.
Causal ablation identifies head L0H3 as the critical selector-routing component: zeroing it mid-transition increases loss by $1.72\nats$ at $K{=}20$, while the next head contributes only $1.07\nats$ (Appendix~\ref{app:ablation}).
Caveat: cascade timing was measured at fixed $n_b$, so the apparent $K$-dependence of lead times may reflect $D$-dependence.

\paragraph{Saddle-point geometry.}
Hessian tracking reveals $\lambda_\text{min} < 0$ during the plateau (saddle, not minimum).
At $K{=}20$, plateau eigenvalues are $\lambda_\text{max} \approx 2.8$ and $\lambda_\text{min} \approx -0.005$, giving $|\lambda_\text{max}/\lambda_\text{min}| \approx 560$.
Across $K$ values, the mean plateau anisotropy ranges from $10{,}800$ to $17{,}700$ (Appendix~\ref{app:hessian}), reflecting variation in $\lambda_\text{min}$ across checkpoints; the escape direction is consistently $500$--$1{,}000\times$ shallower than the dominant curvature.
Consecutive weight displacements have cosine similarity $\approx 0.04$ during the plateau (random walk), snapping to ${\sim}0.8$ at $\tau$ (Figure~\ref{fig:cascade}b).
At convergence, $\lambda_\text{min}$ crosses to positive: the first genuine local minimum.
The optimizer diffuses on a flat saddle until it aligns with the shallow escape direction; noise makes this alignment harder, explaining the entropic stabilization (\S\ref{sec:entropic}).
Both noise-curvature alignment~\citep{wang2023noise} and first-exit-time scaling~\citep{nguyen2019exit} predict that shallower saddles are more sensitive to noise-induced stabilization; the extreme anisotropy here places the system firmly in that regime.

\section{Directional Asymmetry}
\label{sec:asymmetry}

The ``backward'' task $(B, z) \to A$ has $K$-fold ambiguity resolved by $z$.
The ``forward'' task $A \to B$ is unambiguous but requires flat memorization with no shared group structure.
The forward direction is $1.7$--$4.4\times$ slower (Table~\ref{tab:reversal}): the backward task's shared $B$-group structure scaffolds circuit formation, while $A \to B$ requires memorizing each pair independently.
Transfer from forward pre-training helps at high $K$ ($1.31\times$ at $K{=}36$) but hurts at low $K$ ($0.75\times$ at $K{=}5$).
This connects to the reversal curse~\citep{berglund2024reversal}: the information-collapsing direction is slower \emph{because} it lacks the group structure that enables circuit reuse.
Linear networks stay at $\log K$ permanently in both directions, confirming that escape requires nonlinear conditional computation (Appendix~\ref{app:arch}).

\paragraph{Controls and caveats.}
All four task variants at each $K$ use the same surjective map $f$ and the same $(A, B, z)$ triples.
Input sequence lengths differ: forward $A \to B$ has $|A|{+}|z|{+}3 = 9$ input tokens; backward $(B,z) \to A$ has $|B|{+}|z|{+}3 = 11$ input tokens.
Output lengths also differ ($|B|{=}6$ for forward, $|A|{=}4$ for backward).
These differences mean absolute $\tau$ comparisons across directions should be interpreted cautiously; the qualitative pattern (forward consistently slower across all four $K$ values, transfer helping at high $K$) is robust.
Directional asymmetry results are single-seed (seed~42); per-condition replication is not available, but consistency across four $K$ values spanning a $7\times$ range provides robustness through conditions.

\begin{table}[t]
\centering
\caption{\textbf{Directional asymmetry.} Unambiguous forward is consistently slower.}
\label{tab:reversal}
\vspace{2pt}
\small
\begin{tabular}{@{}ccccc@{}}
\toprule
$K$ & $\tau_\text{fwd}$ ($A {\to} B$) & $\tau_\text{bwd}$ ($(B,z) {\to} A$) & Ratio & Transfer\\
\midrule
5  & 2{,}650  & 600    & $4.4\times$ & $0.75\times$ \\
10 & 5{,}400  & 1{,}400  & $3.9\times$ & $0.88\times$ \\
20 & 9{,}500  & 3{,}200  & $3.0\times$ & $1.07\times$ \\
36 & 13{,}050 & 7{,}600  & $1.7\times$ & $1.31\times$ \\
\bottomrule
\end{tabular}
\end{table}

\section{Discussion}
\label{sec:discussion}

\paragraph{What we showed.}
Transformers learn marginals before conditionals.
The metastable regime decomposes into height ($\log K$, set by ambiguity) and duration ($f(D)$, set by dataset size).
The marginal is stabilized by gradient noise, the transition is collective, and an internal cascade (head commitment, synergy, loss drop) precedes it.
Seven candidate mechanisms were tested; all falsified or inconclusive (Appendix~\ref{app:falsifications}).

\paragraph{Limitations and open questions.}
The scaling $\tau \propto D^{1.2}$ is empirical (single model scale, AdamW only).
The D-sweep spans a single decade ($3$K--$36$K) and the fixed-$D$ control has only two $D$ values; confirming the exponent over a wider range remains open.
Unless otherwise noted, results are single-seed (seed~42); where multi-seed data is available (Appendix~\ref{app:multiseed}), seed-to-seed CV ranges from $22\%$ to $71\%$.
The connection to language-model asymmetry is structural analogy, not demonstrated equivalence.
Open: why super-linear ($D^{1.2}$, not $D^{1.0}$)? Can entropic stabilization be formalized via an effective potential~\citep{ziyin2024symmetry} or noise-curvature alignment theory~\citep{wang2023noise}? Does the staging generalize to natural-structure tasks~\citep{papadopoulos2024arrows}?

The entropic-stabilization interpretation predicts that SGD noise covariance during the plateau projects predominantly onto high-curvature (non-escape) directions~\citep{wang2023noise}.
Directly measuring this alignment, by estimating the empirical gradient covariance across mini-batches and projecting onto Hessian eigenvectors, would provide a definitive test.
A complementary experiment would add isotropic gradient noise while keeping labels intact, isolating stochastic stabilization from task degradation.

\paragraph{From diagnosis to prescription.}
A natural next step is to test interventions that accelerate the conditional transition: auxiliary attention losses that reward attending to $z$, architectural inductive biases for routing (e.g., cross-attention over the selector position), or curriculum strategies that present low-$D$ subsets before the full dataset.
Such experiments would move from characterizing the metastable regime to engineering around it.

\bibliography{paper_arxiv_v1}

\appendix

\section{Dataset-Size Sweep}
\label{app:ksweep}

Table~\ref{tab:ksweep} reports $\tau$ versus $K$ at fixed $n_b = 1{,}000$ and $\eta = 10^{-3}$.
Since $D = n_b \times K = 1{,}000K$ in this design, varying $K$ simultaneously varies $D$; the sweep is equivalently a sweep of $D$ from $3{,}000$ to $36{,}000$.
A power-law fit $\tau \propto D^{\delta}$ gives $\delta = 1.19$ (95\% CI $[1.14, 1.31]$, $n = 10$).
All runs reach the $\log K$ plateau within a few hundred steps and subsequently transition to near-zero loss, confirming convergence across the full range.

\begin{table}[h]
\centering
\caption{\textbf{K-sweep at $n_b = 1{,}000$, equivalently a D-sweep.} Since $D = n_b \times K = 1{,}000K$ and $K$ has no independent effect on $\tau$ (\S\ref{sec:duration}), this table is interpretable as $\tau$ vs.\ $D$. Power-law fit: $\tau \propto D^{1.19}$ ($R^2 = 0.98$, 95\% CI $[1.14, 1.31]$).}
\label{tab:ksweep}
\vspace{4pt}
\begin{tabular}{@{}lccccccccccc@{}}
\toprule
$K$   & 3   & 5   & 7     & 10    & 13    & 17    & 20    & 25    & 30    & 36 \\
$D$   & 3K  & 5K  & 7K    & 10K   & 13K   & 17K   & 20K   & 25K   & 30K   & 36K \\
\midrule
$\tau$ & 450 & 800 & 1{,}050 & 1{,}850 & 2{,}100 & 3{,}300 & 3{,}950 & 5{,}250 & 6{,}950 & 8{,}750 \\
\bottomrule
\end{tabular}
\end{table}

\begin{figure}[h]
  \centering
  \includegraphics[width=0.6\textwidth]{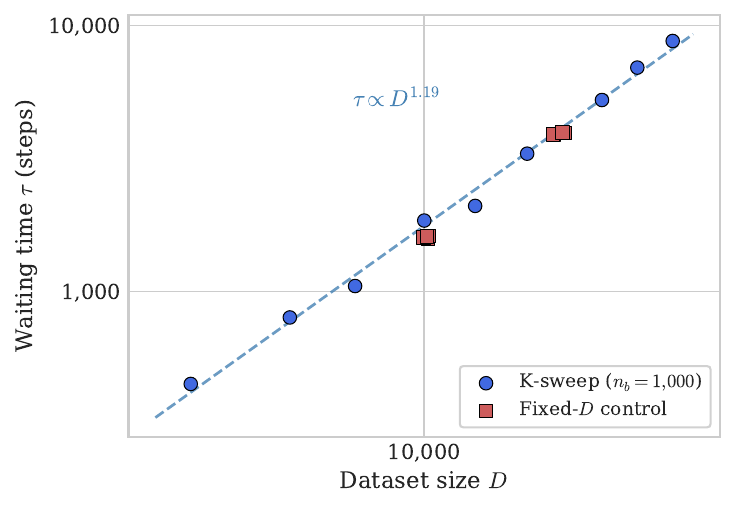}
  \caption{\textbf{$\tau$ versus $D$ on log-log axes.} Blue circles: 10-point K-sweep at $n_b = 1{,}000$, reinterpreted as a D-sweep ($D = 1{,}000K$). Red squares: fixed-$D$ control ($D = 10$K and $20$K, four $K$ values each), confirming $K$-independence. Dashed line: power-law fit $\tau \propto D^{1.19}$.}
  \label{fig:tau-vs-D}
\end{figure}

\section{Fixed-$D$ Control}
\label{app:confound}

\begin{figure}[h]
  \centering
  \includegraphics[width=0.6\textwidth]{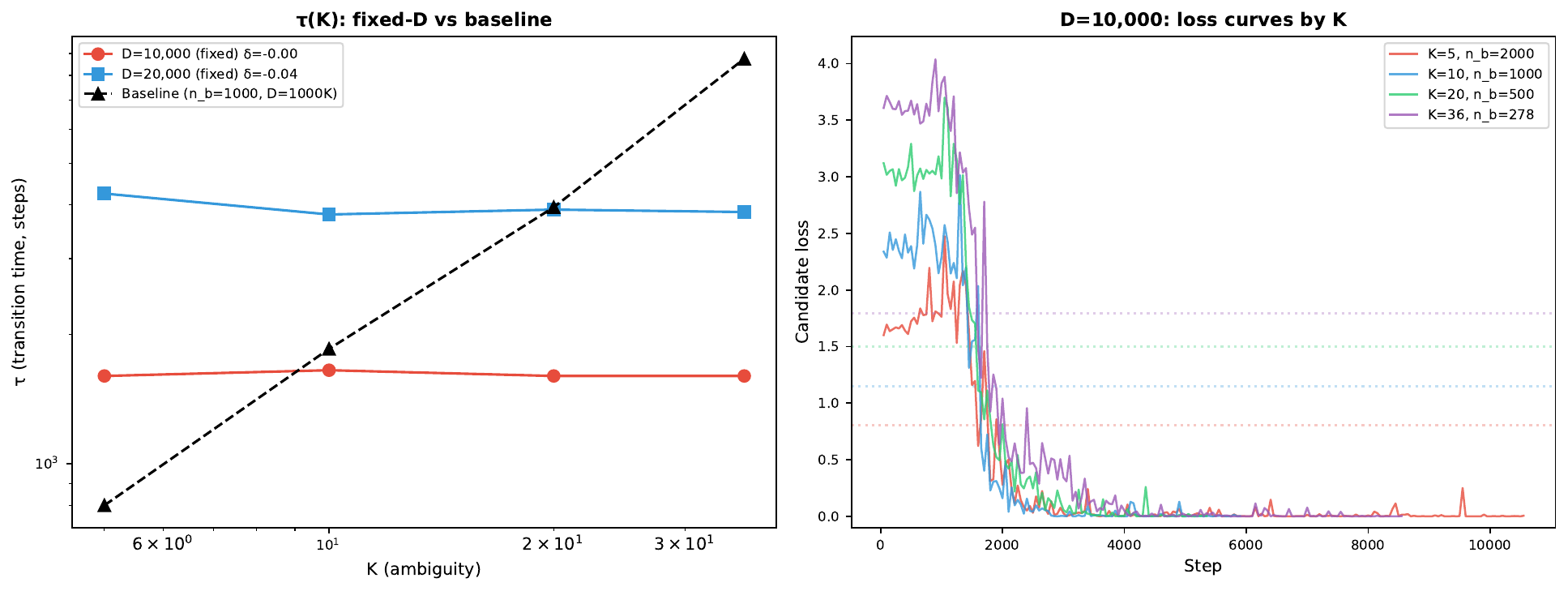}
  \caption{\textbf{Duration depends on $D$, not $K$.} At fixed $D$, $\tau$ is flat across $K$. Doubling $D$ roughly doubles $\tau$.}
  \label{fig:confound}
\end{figure}

Figure~\ref{fig:confound} shows that at fixed dataset size $D$, the waiting time $\tau$ is flat across $K$.
At $D = 10{,}000$, $\tau \approx 1{,}600$ for all $K \in \{5, 10, 20, 36\}$ ($\delta = -0.004$).
Doubling the dataset to $D = 20{,}000$ roughly doubles $\tau$ to ${\approx}3{,}950$, with no $K$-dependence ($\delta = -0.04$, CI $[-0.16, 0.04]$).

\section{Multi-Seed Variability}
\label{app:multiseed}

To assess seed-to-seed variability, we draw on phase-boundary experiments that ran 3 independent seeds (7, 42, 123) at each $(K, \eta)$ condition.
Table~\ref{tab:multiseed} reports all conditions where all 3 seeds converged within 50{,}000 steps.
Since $D = 1{,}000K$, each row is equivalently a point on the $\tau$ vs.\ $D$ curve at a given $\eta$.

The coefficient of variation (CV) ranges from $22\%$ to $71\%$, with a median of $31\%$.
This substantial seed variability is expected for a saddle-escape process: the time to align with a shallow escape direction is inherently stochastic.
Despite this variability, at every $\eta$ where we have two or more $D$ values, $\tau$ increases with $D$: the $D$-scaling is robust across seeds, even though individual $\tau$ values fluctuate.

At the baseline $\eta = 10^{-3}$, the single available multi-seed condition ($K = 36$, $D = 36{,}000$) gives $\tau = 10{,}867 \pm 3{,}349$ (mean $\pm$ std), consistent with the single-seed value of $8{,}750$ from the main $D$-sweep.
The main 11-point $D$-sweep (Appendix~\ref{app:ksweep}) uses seed 42; the phase-boundary data confirms that this seed produces $\tau$ values within the multi-seed distribution.

\begin{table}[h]
\centering
\caption{\textbf{Multi-seed $\tau$ across $D$ values.} 3 seeds per condition from phase-boundary experiments. SE $= \text{std}/\sqrt{3}$.}
\label{tab:multiseed}
\vspace{4pt}
\begin{tabular}{@{}cccccc@{}}
\toprule
$D$ & $K$ & $\eta$ & Mean $\tau$ & Std $\tau$ & CV \\
\midrule
5{,}000  & 5  & $5 \times 10^{-3}$  & 2{,}850  & 1{,}727 & 61\% \\
5{,}000  & 5  & $7 \times 10^{-3}$  & 3{,}250  & 2{,}295 & 71\% \\
5{,}000  & 5  & $1 \times 10^{-2}$  & 2{,}183  & 504     & 23\% \\
10{,}000 & 10 & $3 \times 10^{-3}$  & 5{,}950  & 2{,}299 & 39\% \\
10{,}000 & 10 & $5 \times 10^{-3}$  & 9{,}867  & 3{,}838 & 39\% \\
10{,}000 & 10 & $7 \times 10^{-3}$  & 8{,}533  & 1{,}893 & 22\% \\
20{,}000 & 20 & $2 \times 10^{-3}$  & 5{,}750  & 1{,}316 & 23\% \\
20{,}000 & 20 & $3 \times 10^{-3}$  & 30{,}300 & 8{,}497 & 28\% \\
36{,}000 & 36 & $1 \times 10^{-3}$  & 10{,}867 & 3{,}349 & 31\% \\
\bottomrule
\end{tabular}
\end{table}

\section{Threshold Sensitivity}
\label{app:threshold}

The waiting time $\tau$ is defined as the first step where loss drops below a fraction $\alpha$ of $\log K$.
We test robustness of the power-law exponent $\delta$ to the choice of $\alpha$ across the range $\{0.3, 0.4, 0.5, 0.6, 0.7\}$.
Table~\ref{tab:threshold} reports the fitted exponents.
The exponent varies from $1.12$ ($\alpha = 0.7$) to $1.22$ ($\alpha = 0.3$), with mean $1.18 \pm 0.04$.
This variation is small relative to the confidence interval on any single fit, confirming that the power-law scaling is not an artifact of a particular threshold choice.

\begin{table}[h]
\centering
\caption{\textbf{Threshold sensitivity.} Power-law exponent $\delta$ for $\tau \propto D^\delta$ at different threshold fractions $\alpha$ of $\log K$.}
\label{tab:threshold}
\vspace{4pt}
\begin{tabular}{@{}cccc@{}}
\toprule
$\alpha$ & $\delta$ & 95\% CI & $n$ \\
\midrule
0.3 & 1.22 & $[1.16, 1.34]$ & 10 \\
0.4 & 1.20 & $[1.13, 1.32]$ & 10 \\
0.5 & 1.20 & $[1.14, 1.30]$ & 10 \\
0.6 & 1.17 & $[1.11, 1.31]$ & 10 \\
0.7 & 1.12 & $[1.07, 1.23]$ & 10 \\
\bottomrule
\end{tabular}
\end{table}

\section{Per-Group Evaluation}
\label{app:pergroup}

To test whether the transition is incremental (groups escape one by one) or collective (all groups snap together), we evaluate 200 randomly sampled base groups at multiple checkpoints for $K \in \{10, 20\}$.
For each group, we measure the fraction of its $K$ targets correctly predicted and report the percentage of groups exceeding 80\% and 100\% accuracy.

The key finding is that at $\tau/2$, zero groups exceed 80\% accuracy in either condition.
By $1.5\tau$, nearly all groups have transitioned.
This narrow window confirms a collective mechanism: the selector-routing circuit, once functional, applies to all groups simultaneously rather than being learned group by group.

\begin{table}[h]
\centering
\caption{\textbf{Collective transition.} 200 sampled base groups evaluated at four checkpoints per $K$ value.}
\label{tab:collective}
\vspace{4pt}
\begin{tabular}{@{}cccccc@{}}
\toprule
$K$ & Step & Frac.\ $\tau$ & ${\geq}80\%$ & $100\%$ & Mean acc. \\
\midrule
10 & 500   & 27\% & 0.0\% & 0.0\% & 0.118 \\
10 & 900   & 49\% & 0.0\% & 0.0\% & 0.154 \\
10 & 1{,}800 & 97\% & 31.5\% & 1.5\% & 0.645 \\
10 & 2{,}800 & 151\% & 99.0\% & 75.0\% & 0.969 \\
\midrule
20 & 1{,}900 & 48\% & 0.0\% & 0.0\% & 0.083 \\
20 & 2{,}900 & 73\% & 0.0\% & 0.0\% & 0.192 \\
20 & 3{,}800 & 96\% & 1.5\% & 0.0\% & 0.508 \\
20 & 5{,}900 & 149\% & 87.0\% & 8.5\% & 0.870 \\
\bottomrule
\end{tabular}
\end{table}

\section{Selector Variants}
\label{app:selector}

To verify that the phenomenon is not an artifact of the selector token format, we vary the selector length $|z| \in \{1, 2, 3, 4\}$ at $K = 10$.
Table~\ref{tab:selector} reports the results.
The plateau height remains tightly locked to $\log K$ across all selector lengths (ratio to $\log K$ ranges from $1.007$ to $1.037$).
The waiting time $\tau$ varies modestly from $1{,}450$ to $1{,}850$ steps, with no systematic trend.
The staged-disambiguation phenomenon is robust to the representation of the selector.

\begin{table}[h]
\centering
\caption{\textbf{Selector length variants} at $K = 10$ ($\log K = 2.30\nats$).}
\label{tab:selector}
\vspace{4pt}
\begin{tabular}{@{}cccc@{}}
\toprule
$|z|$ & Plateau (nats) & Plateau / $\log K$ & $\tau$ \\
\midrule
1 & 2.34 & 1.017 & 1{,}650 \\
2 & 2.39 & 1.037 & 1{,}850 \\
3 & 2.33 & 1.014 & 1{,}500 \\
4 & 2.32 & 1.007 & 1{,}450 \\
\bottomrule
\end{tabular}
\end{table}

\section{Architecture Comparison}
\label{app:arch}

We test three nonlinear architectures and one linear baseline.
All nonlinear architectures exhibit the $\log K$ plateau and subsequent sharp transition; the quantitative details differ.

\textbf{Nonlinear architectures.}
Transformers ($\delta = 1.70$, $R^2 = 0.97$), Gated MLPs ($\delta = 1.62$, $R^2 = 0.97$), and RNNs ($\delta = 1.47$, $R^2 = 0.98$) all show power-law scaling of $\tau$ with $D$.
The exponents differ across architectures but all exceed 1, suggesting the super-linear scaling is a property of the optimization landscape rather than a specific architecture.
Note: these runs used $n_b = 1{,}000$ (so $D = 1{,}000K$) and did not control the $D$/$K$ confound; the exponent values should be interpreted with caution.

\textbf{Two-layer linear networks.}
A two-layer linear network ($d = 128$, same tokenizer and dataset) stays at $\log K$ permanently with zero $z$-shuffle gap up to $30{,}000$ training steps.
This confirms that the transition requires nonlinear conditional computation: the marginal solution $P(A \mid B)$ is representable in linear models, but the conditional $P(A \mid B, z)$ requires nonlinear interaction between $B$ and $z$.

\section{Phase Boundary}
\label{app:boundary}

For each $K \in \{5, 10, 20, 36\}$, we sweep over learning rates with 3 seeds per $(K, \eta)$ pair.
A learning rate is classified as ``succeeding'' if all 3 seeds converge within $50{,}000$ steps, and ``failing'' if any seed fails to converge.
The critical learning rate $\eta^*(K)$ is defined as the geometric mean of the largest all-succeed and smallest any-fail values.

Table~\ref{tab:boundary} reports the boundary.
A power-law fit gives $\eta^*(K) = 0.048 \cdot K^{-0.83}$ ($R^2 = 0.96$, $\delta = 0.83 \pm 0.16$).
Near the boundary, $\tau$ diverges: at $K = 36$, $\eta = 2 \times 10^{-3}$, $\tau = 68{,}350$ (i.e., $6.3\times$ the baseline at $\eta = 10^{-3}$).
Across all 24 near-boundary runs that did converge, zero reversions were observed: once the transition initiates, it runs to completion.
Note: these runs used $n_b = 1{,}000$, so $D = 1{,}000K$; the boundary may be $\eta^*(D)$ rather than $\eta^*(K)$.

\begin{table}[h]
\centering
\caption{\textbf{Phase boundary.} Critical learning rate $\eta^*$ per $K$.}
\label{tab:boundary}
\vspace{4pt}
\begin{tabular}{@{}ccccc@{}}
\toprule
$K$ & $D$ & Max all-succeed $\eta$ & Min any-fail $\eta$ & $\eta^*$ \\
\midrule
5  & 5{,}000   & $1.0 \times 10^{-2}$ & $1.5 \times 10^{-2}$ & $1.2 \times 10^{-2}$ \\
10 & 10{,}000  & $7.0 \times 10^{-3}$ & $1.0 \times 10^{-2}$ & $8.4 \times 10^{-3}$ \\
20 & 20{,}000  & $3.0 \times 10^{-3}$ & $5.0 \times 10^{-3}$ & $3.9 \times 10^{-3}$ \\
36 & 36{,}000  & $1.0 \times 10^{-3}$ & $2.0 \times 10^{-3}$ & $1.4 \times 10^{-3}$ \\
\bottomrule
\end{tabular}
\end{table}

\section{Head Ablation}
\label{app:ablation}

We perform zero-ablation (replacing a head's output with zeros) for all 16 heads (4 layers $\times$ 4 heads) at three phases: pre-transition, mid-transition, and post-transition.
Table~\ref{tab:ablation} reports the top heads by loss increase at each phase for $K = 10$ and $K = 20$.

Pre-transition, no single head has an outsized effect (max $\Delta\mathcal{L} < 0.02\nats$).
Mid-transition, head L0H3 emerges as the dominant contributor: ablating it at $K = 10$ increases loss by $1.41\nats$, and at $K = 20$ by $1.72\nats$.
The remaining layer-0 heads also contribute substantially ($1.0$--$1.3\nats$ each), while layer-1 heads contribute an order of magnitude less ($0.1$--$0.14\nats$).
Post-transition, all layer-0 heads become critical ($3.1$--$3.7\nats$), reflecting their role in reading the input tokens.

The concentration of mid-transition effects in L0H3 is consistent with it being the first head to specialize in routing selector information.
Layer-1 heads likely perform downstream computation on the features that layer-0 heads extract.
Caveat: zero-ablation creates out-of-distribution inputs for downstream layers; mean-replacement or rescaling ablations would provide more faithful causal estimates and are left to future work.

\begin{table}[h]
\centering
\caption{\textbf{Head ablation.} Loss increase ($\Delta\mathcal{L}$, nats) when zeroing each head.  Top 4 heads shown per phase.}
\label{tab:ablation}
\vspace{4pt}
\small
\begin{tabular}{@{}clcclc@{}}
\toprule
& \multicolumn{2}{c}{$K = 10$} && \multicolumn{2}{c}{$K = 20$} \\
\cmidrule{2-3} \cmidrule{5-6}
Phase & Head & $\Delta\mathcal{L}$ && Head & $\Delta\mathcal{L}$ \\
\midrule
\multirow{2}{*}{Pre}
& L0H3 & 0.019 && L0H3 & 0.022 \\
& L0H1 & 0.012 && L0H0 & 0.015 \\
\midrule
\multirow{4}{*}{Mid}
& L0H3 & 1.41 && L0H3 & 1.72 \\
& L0H0 & 1.30 && L0H0 & 1.07 \\
& L0H2 & 1.28 && L0H2 & 1.05 \\
& L0H1 & 1.27 && L0H1 & 1.03 \\
\midrule
\multirow{4}{*}{Post}
& L0H3 & 3.65 && L0H3 & 3.69 \\
& L0H2 & 3.37 && L0H0 & 3.61 \\
& L0H1 & 3.36 && L0H2 & 3.41 \\
& L0H0 & 3.13 && L0H1 & 3.28 \\
\bottomrule
\end{tabular}
\end{table}

\section{Hessian Details}
\label{app:hessian}

We compute the largest and smallest eigenvalues of the Hessian via power iteration (50 iterations, batch size 512) at checkpoints spanning the plateau and transition for $K \in \{10, 20, 36\}$.

During the plateau, the loss landscape is a saddle: $\lambda_\text{min} < 0$ for all $K$ values.
At $K = 20$, representative plateau eigenvalues are $\lambda_\text{max} \approx 2.8$ and $\lambda_\text{min} \approx -0.005$, giving a pointwise ratio $|\lambda_\text{max}/\lambda_\text{min}| \approx 560$.
However, $\lambda_\text{min}$ varies across checkpoints within the plateau; the mean anisotropy (averaged over plateau checkpoints) is $12{,}100$ at $K{=}20$ and ranges from $10{,}800$ to $17{,}700$ across $K$ values (Table~\ref{tab:hessian_anisotropy}).
The escape direction is consistently $500$--$1{,}000\times$ shallower than the dominant curvature.

At the transition, $\lambda_\text{max}$ spikes to $30$--$80$, reflecting the rapid weight changes as the model reorganizes.
Post-convergence, $\lambda_\text{min}$ crosses to a small positive value ($\approx +0.0001$): the first genuine local minimum encountered during training.

The anisotropy does not scale systematically with $K$ ($\gamma = 0.06$, $R^2 = 0.05$), consistent with the falsification of the ``curvature scales with $K$'' hypothesis.

\textbf{Robustness checks.}
We recomputed $\lambda_\text{max}$ and $\lambda_\text{min}$ at $K = 20$ across batch sizes $\{256, 512, 1024, 2048\}$ and iteration counts $\{25, 50, 100, 200\}$.
The coefficient of variation is $< 0.1$ for $\lambda_\text{max}$ and $< 0.15$ for $\lambda_\text{min}$.
The sign of $\lambda_\text{min}$ (negative on plateau, positive post-convergence) is consistent across all settings.

\begin{table}[h]
\centering
\caption{\textbf{Hessian anisotropy} ($|\lambda_\text{max}/\lambda_\text{min}|$) during the plateau for each $K$.}
\label{tab:hessian_anisotropy}
\vspace{4pt}
\begin{tabular}{@{}ccc@{}}
\toprule
$K$ & Anisotropy (mean) & $\lambda_\text{max}$ \\
\midrule
3 & $17{,}700$ & 9.83 \\
5 & $11{,}700$ & 5.90 \\
10 & $10{,}800$ & 4.59 \\
20 & $12{,}100$ & 2.83 \\
36 & $17{,}600$ & 0.95 \\
\bottomrule
\end{tabular}
\end{table}

\section{Hierarchical Task}
\label{app:hierarchical}

We tested a two-level hierarchical version of the task to see whether the model would produce a staircase of plateaus (one per disambiguation level).
In this variant, each base string $B$ maps to $K_1$ groups, each of which maps to $K_2$ targets, yielding $K_1 \times K_2$ total targets per $B$.
Two selectors $z_1$ and $z_2$ resolve the ambiguity at each level.

\textbf{Small hierarchy ($K_1 = 5$, $K_2 = 4$, $D = 20{,}000$):} The model plateaus at $\log(20) \approx 3.0\nats$ and transitions directly to near-zero loss.
No intermediate plateau appears at $\log(5) \approx 1.6\nats$.
The model does not learn to resolve one level of ambiguity before the other.

\textbf{Large hierarchy ($K_1 = 20$, $K_2 = 10$, $D = 200{,}000$):} The model remains stuck at $\log(200) \approx 5.3\nats$ for $100{,}000$ steps, likely due to the large dataset size.
No intermediate plateau at $\log(20)$ is observed during this window.

Both selectors provide gradient signal from step one; the staging we observe in the main task is between marginal and conditional prediction, not between hierarchical levels of ambiguity.
This negative result suggests the ``marginals before conditionals'' principle does not straightforwardly extend to hierarchical disambiguation.

\section{Gradient Dissipation}
\label{app:qslope}

We measure the total gradient dissipation $Q = \sum_{t \in \text{transition}} \eta \|\nabla \mathcal{L}\|^2$ during the transition window (from $\tau/2$ to $2\tau$).
At $\eta = 10^{-3}$, $Q$ scales linearly with $\log K$ (slope $= 4.29$, $R^2 = 0.991$; Table~\ref{tab:qslope}).
At $\eta = 3 \times 10^{-4}$, the slope increases to $9.91$ ($R^2 = 0.913$), and at $\eta = 3 \times 10^{-3}$ the relationship degrades (slope $= 110.2$, $R^2 = 0.732$).

The $\log K$ scaling of $Q$ at moderate learning rates suggests that the optimizer must dissipate a quantity of gradient energy proportional to the information-theoretic gap $\log K$ during the transition.
The strong $\eta$-dependence of the slope is expected: larger steps dissipate more energy per update.
Caution: these runs used fixed $n_b = 1{,}000$, so the $\log K$ dependence may partly reflect $D$-dependence.

\begin{table}[h]
\centering
\caption{\textbf{Gradient dissipation $Q$} during transition, by $K$ and $\eta$.}
\label{tab:qslope}
\vspace{4pt}
\begin{tabular}{@{}cccc@{}}
\toprule
$K$ & $Q$ ($\eta = 3 \times 10^{-4}$) & $Q$ ($\eta = 10^{-3}$) & $Q$ ($\eta = 3 \times 10^{-3}$) \\
\midrule
5  & 2.72 & 3.96 & 7.05 \\
10 & 5.92 & 7.42 & 3.48 \\
20 & 16.4 & 9.90 & 159.8 \\
\bottomrule
\end{tabular}
\end{table}

\section{Falsification Summary}
\label{app:falsifications}

We tested seven candidate mechanisms for the plateau.
Table~\ref{tab:falsifications} summarizes each hypothesis, its prediction, the experimental result, and the verdict.
Six are clearly falsified; one is inconclusive due to numerical instability.

\begin{table}[h]
\centering
\caption{\textbf{Tested mechanisms and their outcomes.}}
\label{tab:falsifications}
\vspace{4pt}
\small
\begin{tabular}{@{}p{3cm}p{3.5cm}p{3.5cm}l@{}}
\toprule
Hypothesis & Prediction & Result & Verdict \\
\midrule
Gradient cancellation drives plateau & $\tau \propto$ number of groups $G$ & $\beta \approx 0.02$; no scaling with $G$ & Falsified \\
\addlinespace
Barrier crossing (noise helps) & More noise $\to$ faster escape & LR: $3.6\times$ slower; BS: $1.8\times$ (tokens) & Falsified \\
\addlinespace
Incremental group coverage & Groups solved independently over time & 0\% of groups ${\geq}80\%$ at $\tau/2$ & Falsified \\
\addlinespace
Label noise breaks symmetry & Noise perturbs away from flat solution & $16\times$ delay ($p{=}0.2$); confounded$^*$ & Falsified$^*$ \\
\addlinespace
$K$ drives duration & $\tau \propto K^\delta$ at fixed $D$ & $\delta = -0.004$; no $K$ effect & Falsified \\
\addlinespace
Linear networks suffice & Two-layer linear escapes & Stays at $\log K$ permanently & Falsified \\
\addlinespace
Curvature scales with $K$ & $|\lambda_\text{max}/\lambda_\text{min}| \propto K$ & Anisotropy $\sim 10^4$ for all $K$ & Inconclusive \\
\bottomrule
\multicolumn{4}{@{}l@{}}{\scriptsize $^*$Label noise also degrades $I(A; z \mid B)$; the $16\times$ delay conflates noise stabilization with task degradation (\S\ref{sec:entropic}).}
\end{tabular}
\end{table}

\end{document}